\documentclass[11pt]{article}

\usepackage[preprint]{acl}

\usepackage{times}
\usepackage{latexsym}

\usepackage[T1]{fontenc}

\usepackage[utf8]{inputenc}

\usepackage{microtype}

\usepackage{inconsolata}

\usepackage{graphicx}

\usepackage{multirow}
\usepackage[table,xcdraw]{xcolor}
\usepackage{colortbl}

\usepackage{amsmath}
\usepackage{subcaption}

\title{Evaluation of Automatic Speech Recognition Using Generative Large Language Models}

\author{
  Thibault Bañeras-Roux\textsuperscript{1},
  Shashi Kumar\textsuperscript{1,2},
  Driss Khalil\textsuperscript{1},
  Sergio Burdisso\textsuperscript{1}, \\
  \textbf{Petr Motlicek\textsuperscript{1,3}},
  \textbf{Shiran Liu\textsuperscript{1}},
  \textbf{Mickael Rouvier\textsuperscript{4}},
  \textbf{Jane Wottawa\textsuperscript{5}},
  \textbf{Richard Dufour\textsuperscript{6}} \\
  \\
  \textsuperscript{1}Idiap Research Institute, Martigny, Switzerland \\
  \textsuperscript{2}EPFL, Lausanne, Switzerland \\
  \textsuperscript{3}Brno University of Technology, Czech Republic \\
  \textsuperscript{4}Avignon University, Avignon, France \\
  \textsuperscript{5}Le Mans University, Le Mans, France \\
  \textsuperscript{6}Nantes University, Nantes, France
  \\
  \small{
    \textbf{Correspondence:} \href{mailto:thibault.roux@idiap.ch}{thibault.roux@idiap.ch}
  }
}

\begin{document}
\maketitle
\begin{abstract}
Automatic Speech Recognition (ASR) is traditionally evaluated using Word Error Rate (WER), a metric that is insensitive to meaning. Embedding-based semantic metrics are better correlated with human perception, but decoder-based Large Language Models (LLMs) remain underexplored for this task. This paper evaluates their relevance through three approaches: (1) selecting the best hypothesis between two candidates, (2) computing semantic distance using generative embeddings, and (3) qualitative classification of errors. On the HATS dataset, the best LLMs achieve 92--94\% agreement with human annotators for hypothesis selection, compared to 63\% for WER, also outperforming semantic metrics. Embeddings from decoder-based LLMs show performance comparable to encoder models. Finally, LLMs offer a promising direction for interpretable and semantic ASR evaluation.
\end{abstract}

\section{Introduction}
 
 
 

Automatic Speech Recognition (ASR) is a fundamental technology for voice control applications, transcription, and accessibility. Reliable evaluation of ASR systems is essential for measuring progress and guiding model improvement. Historically, Word Error Rate (WER) has dominated this landscape, but this metric has well-known limitations: it is rigid and case-sensitive, not necessarily capturing information important for the task or for human perception.
 
Facing these challenges, several alternatives have been proposed. Semantic metrics~\cite{kim2021semantic, zhangbertscore} based on word embeddings have shown promising performance, particularly those exploiting contextual encoder models like BERT~\cite{devlin2019bert}. These approaches better capture the semantic nuances of language, explaining their better correlation with human judgments. However, while encoders have been frequently explored for this task~\cite{roux2022qualitative, kim2021semantic, le2016better, gordeeva2021meaning}, decoder-based Large Language Models (LLMs) -- such as GPT~\cite{radford2018improving}, Llama~\cite{touvron2023llama}, or Gemma~\cite{team2024gemma} -- have received little attention. Yet these models have demonstrated remarkable and versatile capabilities across a wide range of tasks~\cite{ahuja2023mega, labrak2024biomistral, naveed2025comprehensive}, suggesting they could offer innovative perspectives for ASR evaluation.
 
This study explores precisely the capabilities of decoder-based LLMs in the specific context of evaluating automatic speech recognition systems, along two complementary axes. The first axis focuses on the ``LLM as judge'' paradigm: can we leverage the deep contextual and semantic understanding of LLMs to directly evaluate the quality of transcription hypotheses? The second axis examines the quality of vector representations extracted from these models: how do embeddings from decoder LLMs compare to those from encoders in capturing relevant semantic similarities?
 
Our experiments rely on the HATS dataset~\cite{baneras2023hats}, which provides human annotations of speech recognition errors and has shown that semantic metrics correlate significantly better with human perception than WER. We evaluate (1) the capacity of LLMs to select the best hypothesis among two candidates, (2) the effectiveness of different pooling strategies applied to LLM embeddings to build a semantic similarity metric, and (3) the capacity of LLMs to assign qualitative labels to transcription errors. Our results reveal that the best LLMs not only outperform WER and CER, but also sophisticated semantic metrics based on encoders, while offering better interpretability of errors.

\section{Related Work}
\label{sec:sota}
 
 
 

\subsection{Limitations of Word Error Rate}
 
As mentioned previously, Word Error Rate has been the reference metric for evaluating automatic speech recognition systems. Yet WER presents significant limitations well-documented in the literature~\cite{wang2003word, morris2004and, he2011word}. First, this metric operates at the lexical level without any consideration of semantic context, treating errors that severely affect comprehension identically to minor errors with no impact on meaning. Second, WER does not necessarily correlate with downstream tasks~\cite{favre2013automatic} that depend on the produced transcriptions, suggesting that WER improvement does not guarantee perceptible improvement for the end user. Finally, several studies have demonstrated poor correlation between WER and human perception of transcription quality~\cite{kim2022evaluating, thennal2025advocating}, questioning its usefulness for evaluating the real performance of ASR systems intended for human reading.
 
These limitations have important practical implications. Indeed, studies have shown that the ranking of ASR systems can differ significantly depending on the metric used for evaluation~\cite{baneras2024comprehensive}. For a specific objective -- for example, generating subtitles to be read by humans -- it may be more relevant to use a metric that correlates better with human perception rather than optimizing a WER that does not capture relevant perceptual nuances~\cite{kafle2017evaluating,nam2019simulation}.
 
\subsection{Semantic Metrics Based on Embeddings}
 
 
Facing WER's limitations, several alternatives have been proposed. Among these, semantic metrics based on word embeddings have proven particularly promising. These approaches exploit vector representations of language obtained by word embedding or context models, allowing capture of semantic similarities between reference and transcription hypothesis.
 
A first generation of these metrics exploited static word embeddings to improve ASR evaluation in machine translation context~\cite{le2016better}. More recently, the emergence of powerful contextual models, particularly bidirectional encoders like BERT and its multilingual variants, has enabled development of more sophisticated semantic metrics. These contextual models offer richer representations, sensitive to word position and the context surrounding each linguistic unit. Among notable semantic approaches, the \textit{SemDist} metric~\cite{kim2021semantic} proposes a distance derived from cosine similarity between embeddings of a reference and a hypothesis. This metric has demonstrated significantly better correlation with human perception compared to WER, validating the hypothesis that semantics is a key factor for perceptual evaluation of transcription quality.
 
\citeauthor{mccowan2004use} defines the ideal metric for automatic speech recognition on several criteria including interpretability. The problem is that semantic metrics are primarily constructed from cosine similarities between embeddings, producing scores such as: 0.62, 0.75, or 0.81. Unlike error percentages (as WER and CER do), these scores are poorly interpretable.
 
To address this issue, some work~\cite{gordeeva2021meaning, roy2021semantic} has proposed a semantic metric that can be adapted to a downstream task and could be interpreted as a weighted word error rate based on the semantic and lexical severity of an error. Other work~\cite{baneras2024paradigm} has also proposed paradigms for interpreting ASR errors based on their severity and impact on semantic metrics using contextual embeddings. These approaches offered more granular understanding of errors, allowing identification of which error types most affect perceived transcription quality.
 
 
\subsection{Human Perception and Interpretable Evaluation}
 
A complementary research axis has focused on explicit alignment between automatic metrics and human perception. Several works have proposed evaluation models based on human perception~\cite{itoh2015metric, sasindran2024semascore}.
 
Empirical studies have collected human annotations to evaluate how speakers perceive the quality of automatic transcriptions. These annotations enabled creation of datasets like HATS, a French dataset that integrate explicit human judgments on error quality, allowing validation and comparison of different evaluation metrics. This study showed greater correlation with human perception than WER, confirming another study~\cite{kim2022evaluating} conducted on private datasets integrating human perception. Another recent study~\cite{thennal2025advocating} also demonstrated using human annotations that character error rate is more relevant than WER in a multilingual setting.

\section{Experiments}
 
 
 
 
We rely on results reported in the HATS dataset to establish a reference point for metric evaluation. The table~\ref{tab:hats-original-metrics} presents the percentage of agreement between different automatic metrics and human judgments.
 
To better interpret these results, three subsets of the corpus are considered, constructed based on the level of agreement between human annotators: (i) examples with strict 100\% agreement, (ii) those with at least 70\% agreement, and (iii) the complete dataset. This split allows analysis of metric behavior based on annotation reliability: strong agreement is presumed to reflect more "obvious" cases, while lower agreement corresponds to more ambiguous examples or where annotators might have chosen randomly.
 
Agreement columns indicate the proportion of cases where the metric is consistent with human preference, while Equal columns correspond to situations where the metric cannot distinguish between outputs (which does not occur for cosine similarity-based metrics as these are continuous values). Agreement is calculated as follows:
\[
\text{Agreement} = \frac{\max(A, B)}{A + B}
\]
 
Given BERTScore's weaker performance, we focus in this study on SemDist. For our experiments, we use the SDialog toolkit~\cite{burdisso2026sdialog}.
 
\begin{table*}[htb]
\centering
\begin{tabular}{lrrrrrr}
\hline
 & \multicolumn{2}{c}{\textbf{$=$ 100 \%}} & \multicolumn{2}{c}{\textbf{$\geq$ 70 \%}} & \multicolumn{2}{c}{\textbf{Full}} \\
\multirow{-2}{*}{\textbf{Metric}} & \multicolumn{1}{l}{Agreement} & \multicolumn{1}{l}{Equal} & \multicolumn{1}{l}{Agreement} & \multicolumn{1}{l}{Equal} & \multicolumn{1}{l}{Agreement} & \multicolumn{1}{l}{Equal} \\ \hline
Word Error Rate & 63 & {\color[HTML]{9B9B9B} 23} & 53 & {\color[HTML]{9B9B9B} 28} & 49 & {\color[HTML]{9B9B9B} 28} \\
Character Error Rate & 77 & {\color[HTML]{9B9B9B} 17} & 64 & {\color[HTML]{9B9B9B} 21} & 60 & {\color[HTML]{9B9B9B} 22} \\
BERTScore CamemBERT-large & 80 & {\color[HTML]{9B9B9B} 0} & 68 & {\color[HTML]{9B9B9B} 0} & 65 & {\color[HTML]{9B9B9B} 0} \\
SemDist CamemBERT-large & 80 & {\color[HTML]{9B9B9B} 0} & 71 & {\color[HTML]{9B9B9B} 0} & 67 & {\color[HTML]{9B9B9B} 0} \\
SemDist Sentence CamemBERT-large & 90 & {\color[HTML]{9B9B9B} 0} & 78 & {\color[HTML]{9B9B9B} 0} & 73 & {\color[HTML]{9B9B9B} 0} \\ \hline
\end{tabular}%
\caption{Percentage of agreement between metrics and human perception, based on annotator agreement level. The mentions \textbf{100\%} and \textbf{70\%} designate subsets of the HATS dataset based on percentage of agreement between annotators.}
\label{tab:hats-original-metrics}
\end{table*}

\subsection{Selecting the Best Hypothesis Using an LLM}
\label{sec:hypA-or-hypB}
 
 
 
We use different LLMs to determine, among two transcription hypotheses, which one is closest to a given reference. This experiment aims to evaluate the capacity of different LLMs to compare two incorrect hypotheses and identify the one that best matches the reference transcription, considering both lexical fidelity and semantic coherence.
 
The model is queried via a one-shot prompt, comprising an annotated example followed by a case to predict. The model is explicitly asked to justify its answer, then conclude with a decision in the form of \texttt{A} or \texttt{B}. This formulation allows exploitation of the model's reasoning (chain-of-thought) while facilitating automatic extraction of the final prediction.
 
The prompt used is as follows (translated from the original):
\small
\begin{quote}
\texttt{\textbf{User:} A reference is an exact transcription of audio. Two incorrect hypotheses are proposed.
Explain your reasoning and end your sentence by writing 'A', 'B', 'a', or 'b' if undecided. \\ Reference: it's to him self \\
Hypothesis A: are to him self \\
Hypothesis B: it's uh to him-self} \\ \\
\texttt{\textbf{Assistant:} Although hypothesis B contains a disfluency ('uh'), it corresponds much better to the reference in terms of words and meaning. The disfluency can be tolerated if it is part of the original, while hypothesis A's grammatical error is more problematic. Therefore, the most acceptable transcription is hypothesis B.} \\ \\
\texttt{\textbf{User:} Reference: <reference> \\
Hypothesis A: <hypothesis A> \\
Hypothesis B: <hypothesis B> \\}
\end{quote}
\normalsize

The results presented in Table~\ref{tab:correlation-hypA-or-hypB} reveal that generative LLMs are capable of efficiently selecting the best transcription hypothesis among two candidates, in agreement with human perception. A general progression of performance is observed with model size and generation, although size is not the only determining criterion: thus, \texttt{qwen3-8B}, though noticeably smaller, outperforms \texttt{gemma3-27b}, and \texttt{qwen3.5-27b} surpasses \texttt{qwen3-30b} despite comparable parameter counts. The most recent models, \texttt{gpt-4.1} and \texttt{qwen3.5-35b}, achieve the best performance with 94\% and 92\% agreement with annotators, respectively, on the full agreement subset. It is worth noting that an open-access model like \texttt{qwen3.5-35b} shows performance comparable to \texttt{gpt-4.1}, suggesting that accessible alternatives can rival state-of-the-art proprietary models for this task.
 
\begin{table}[htb]
\centering
\begin{tabular}{lrrr}
\hline
\textbf{LLM} & \textbf{$=$ 100 \%} & \textbf{$\geq$ 70 \%} & \textbf{Full} \\ \hline
GPT-4o & 92 & 83 & 78 \\
GPT-4.1 & \textbf{94} & \textbf{85} & \textbf{79} \\ \hline
gemma3-27b & 72 & 63 & 61 \\
gemma4-31b & 87 & 78 & 73 \\
Qwen3-0.6B & 50 & 47 & 47 \\
Qwen3-1.7B & 59 & 58 & 56 \\
Qwen3-8B & 80 & 74 & 72 \\
Qwen3-30B & 84 & 75 & 71 \\
Qwen3.5-27B & 91 & 83 & 77 \\
Qwen3.5-35B & \textbf{92} & \textbf{83} & \textbf{78} \\ \hline
\end{tabular}%
\caption{Percentage of agreement between LLM choices (selecting the best hypothesis) and human perception, based on annotator agreement level.}
\label{tab:correlation-hypA-or-hypB}
\end{table}

Even more remarkably, the performance of the best LLMs surpasses that of classic automatic ASR evaluation metrics such as WER and CER, which operate at the character or word level without semantic consideration, but also that of more sophisticated metrics like SemDist, including in its best configuration using sentence embeddings from CamemBERT-large. This result suggests that generative LLMs, thanks to their deep contextual and semantic understanding of language, succeed in capturing perceptual nuances that traditional metrics do not model, such as tolerance for disfluencies or preference for grammatical coherence and overall utterance meaning. These capabilities also open promising perspectives for automatic annotation of perceptual data, where LLMs could partially substitute or complement human annotation at lower cost. Furthermore, this approach also enables direct comparisons between the performance of two ASR systems, evaluating their outputs relatively rather than absolutely, which is particularly relevant in scenarios where one seeks to deploy the system performing best with respect to human perception.
 
\subsection{Semantic Metric Based on Decoder LLM Embeddings}
\label{sec:semdist}
 
We evaluate the correlation of the \textit{SemDist} metric, defined as a distance derived from cosine similarity between reference and hypothesis embeddings, on the HATS dataset.
 
To obtain representations of transcriptions, we use different families of LLMs, covering multiple sizes and training methods. The text is first tokenized into a token sequence, then transformed by the LLM into an embedding sequence, corresponding to vector representations of each token.
 
 
The \textit{SemDist} metric requiring a fixed-dimension representation, this sequence is aggregated via a \textit{pooling} operation. We compare several pooling strategies, ranging from simple methods like averaging embeddings to weighted approaches.
 
The objective is to evaluate the impact of these choices on representation quality, particularly their ability to capture semantic similarities reflected by human annotations. All combinations between LLM families and pooling methods are thus evaluated in terms of correlation between \textit{SemDist} and HATS dataset annotations.
 
 
 
\paragraph{Pooling strategies.}
We consider the following pooling strategies, where $t_i$ denotes the embedding of the $i$-th token and $n$ denotes the sequence length:
 
\begin{itemize}
    \item \textbf{Last token}: using the last token embedding, $t_n$.
    
    \item \textbf{Second-to-last}: using the embedding of the token preceding the last, $t_{n-1}$.
    
    \item \textbf{Mean}: average of embeddings of all tokens, 
    $\frac{1}{n}\sum_{i=1}^{n} t_i$.
    
    \item \textbf{Mean without last}: average of embeddings excluding the last token, 
    $\frac{1}{n-1}\sum_{i=1}^{n-1} t_i$.
    
    \item \textbf{Weighted mean}: weighted average of embeddings of all tokens, where weights increase linearly with position (i.e., weight $i$ for token $t_i$), 
    $\frac{\sum_{i=1}^{n} i \, t_i}{\sum_{i=1}^{n} i}$.
    
    \item \textbf{Weighted mean without last}: weighted average of embeddings excluding the last token, with the same positional weights, 
    $\frac{\sum_{i=1}^{n-1} i \, t_i}{\sum_{i=1}^{n-1} i}$.
    
    \item \textbf{Mean of last 4 tokens}: average of embeddings of the last four tokens, 
    $\frac{1}{4}\sum_{i=n-3}^{n} t_i$.
\end{itemize}

\begin{table*}[h]
\centering
\small
\begin{tabular}{lrrrrrrr}
\hline
\textbf{Model} & \textbf{Last} & \textbf{Mean} & \textbf{Mean*} & \textbf{Wgt.} & \textbf{Wgt.*} & \textbf{4 Last} & \textbf{2nd Last} \\
\hline
gemma-2-2b & 73 & 79 & 79 & 79 & 78 & 82 & 83 \\
gemma-2b & 76 & 80 & 80 & 79 & 80 & 81 & 80 \\
gemma-3-1b-pt & 75 & 76 & 74 & 77 & 76 & 76 & 75 \\
gemma-3-27b-it & 77 & 83 & 83 & 81 & 82 & 82 & 82 \\
gemma-4-31B & 72 & 75 & 73 & 77 & 76 & 76 & 75 \\
gemma-7b & 65 & 69 & 71 & 71 & 72 & 73 & 80 \\
OLMo-2-1124-7B & 73 & 82 & 80 & 83 & 80 & 83 & 83 \\
Mistral-7B-v0.3 & 79 & 83 & 84 & 84 & 85 & 85 & 85 \\
Qwen3-0.6B & 70 & 80 & 80 & 78 & 78 & 81 & 81 \\
Qwen3-0.6B-Base & 74 & 79 & 79 & 81 & 80 & 80 & 80 \\
Qwen3-1.7B & 72 & 83 & 83 & 80 & 81 & 87 & 87 \\
Qwen3-1.7B-Base & 74 & 82 & 81 & 81 & 82 & 86 & 86 \\
Qwen3-30B & 72 & 84 & 84 & 83 & 82 & 84 & 82 \\
Qwen3-4B & 77 & 85 & 83 & 82 & 81 & 79 & 80 \\
Qwen3-4B-Base & 74 & 82 & 83 & 82 & 82 & 82 & 83 \\
Qwen3-8B & 75 & 85 & 83 & 83 & 83 & 85 & 85 \\
Qwen3-8B-Base & 71 & 78 & 78 & 80 & 77 & 82 & 82 \\
Qwen3-Embedding-0.6B & \textbf{88} & 84 & 85 & 85 & 84 & 83 & 82 \\
Qwen3-Embedding-4B & \textbf{88} & 86 & 85 & 86 & 84 & 85 & 86 \\
Qwen3-Embedding-8B & \textbf{88} & \textbf{89} & \textbf{87} & \textbf{89} & \textbf{87} & \textbf{88} & \textbf{89} \\
Qwen3.5-27B & 59 & 77 & 76 & 74 & 76 & 78 & 77 \\
Qwen3.5-35B & 68 & 68 & 68 & 67 & 68 & 67 & 67 \\  \hline
Mean & \textit{74} & \textit{80} & \textit{80} & \textit{80} & \textit{80} & \textit{81} & \textit{81} \\ \hline
\end{tabular}
\caption{SemDist performance by model and pooling technique on HATS dataset subset with annotator consensus.\\
\small * = without last token | Last = Last token | 2nd Last = Second-to-last}
\label{tab:semdist_comparison}
\end{table*}

Contrary to initial intuitions, the analysis reveals a weak relationship between model size and correlation of embeddings with human perception. While it is common to observe better performance for larger models, we observe little of this phenomenon, sometimes even the reverse: Qwen3.5-35B shows weaker performance than Qwen3.5-27B. 
Interestingly: the LLMs with the best performance for classifying a hypothesis as better (see Table~\ref{tab:correlation-hypA-or-hypB}) do not necessarily have the most relevant embeddings for the SemDist metric (Qwen3.5-27B scores 91\% for hypothesis selection versus 78\% with the mean of 4 last tokens).
 
A notable result is that representations based solely on the last token are, in most cases, surpassed by alternatives. For example, Qwen3.5-27B achieves only 59\% with last token pooling compared to at least 74\% for weighted pooling, the second worst method for this LLM, which has a 15-point gap. We attribute this to the training objective of LLMs: the next-token prediction task optimizes representations to predict future tokens rather than capture overall semantic content. The last token, positioned at sequence end where it must predict an uninformative continuation, carries little semantic information about the complete sentence.
 
Despite its potential ability to erase and lose information, mean pooling proves particularly effective, often equaling or surpassing more sophisticated strategies. We explain this through a geometric property of the embedding space: when the hypothesis and reference have comparable lengths (expected in an ASR task), semantically similar pairs exhibit convergent embedding trajectories, causing their averages to cluster in latent space. 
 
Unsurprisingly, models explicitly fine-tuned for embedding tasks considerably outperform generic LLMs of comparable size, even when using the last token embedding, which was the token used for LLM fine-tuning~\cite{zhang2025qwen3}. Qwen3-Embedding-8B achieves 89\% with mean pooling compared to 85\% for Qwen3-8B, a 4-point improvement. Remarkably, this fine-tuning reduces the gap between the last token and other methods, suggesting that task-specific training reorients the last token to serve as a coherent semantic representation rather than simply as a next-token predictor.
 
 

\subsection{Generative LLM for Classifying Hypotheses}
\label{sec:llm-classify-category}
 
 
 
 
In this section, we evaluate the capacity of language models to classify a hypothesis given a reference. Unlike Section~\ref{sec:hypA-or-hypB}, where the task consisted of selecting the best hypothesis among two candidates for the same reference, we consider here each (reference, hypothesis) pair independently, and the model must assign a qualitative class among four categories. The objective is to perform an intrinsic and objective evaluation of the model.
 
Given a reference and a hypothesis, the model must assign one of the following labels:
\begin{itemize}
    \item \textbf{identical}: the hypothesis is identical to the reference (or differ only in case or hyphenation);
    \item \textbf{useful}: meaning is preserved despite minor errors (normalization, punctuation, spelling, capitals, abbreviations, slight syntactic variations without loss of comprehension);
    \item \textbf{bad}: meaning is partially altered (significant errors on keywords, important substitutions or omissions, but some content remains understandable);
    \item \textbf{incomprehensible}: meaning is completely lost (the sentence cannot be understood or correctly interpreted relative to the reference).
\end{itemize}
 
Each pair is annotated automatically by the considered model, without explicit comparison with other hypotheses. This formulation allows evaluation of the intrinsic capacity of the model to judge hypothesis quality absolutely, rather than relatively.
 
In parallel with this categorical annotation, we compute a SemDist score between reference and hypothesis using the best-performing encoder LLM to our knowledge: \texttt{sentence-camembert-large}. This measure provides an independent quantitative signal, allowing analysis of coherence between the model's qualitative judgments and a continuous metric that simulates human perception.
 
\begin{figure*}[htbp]
\centering
\begin{subfigure}[b]{0.48\linewidth}
    \centering
    \includegraphics[width=\linewidth]{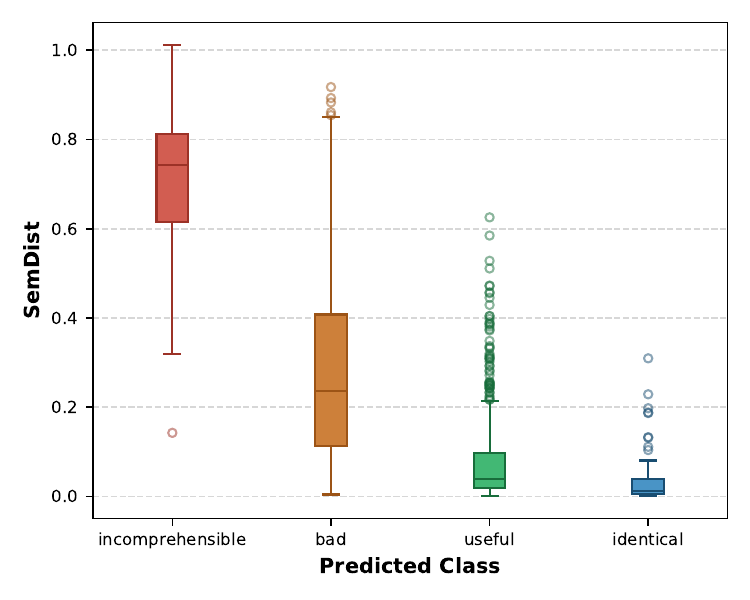}
    \caption{GPT-4.1}
    \label{fig:gpt41}
\end{subfigure}
\hfill
\begin{subfigure}[b]{0.48\linewidth}
    \centering
    \includegraphics[width=\linewidth]{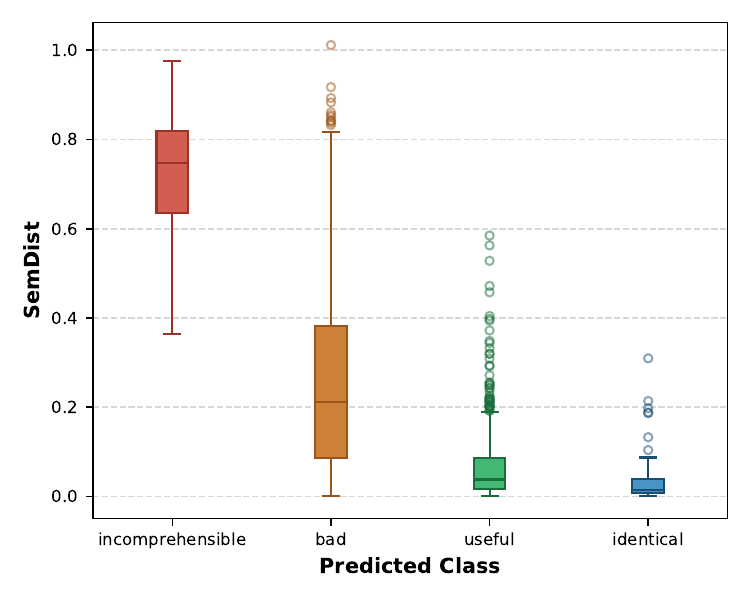}
    \caption{Qwen3.5-35B}
    \label{fig:qwen35}
\end{subfigure}
\caption{Box plot distribution of SemDist scores by class annotated by an LLM.}
\label{fig:model_pooling_performance}
 
\end{figure*}
 
Figure~\ref{fig:model_pooling_performance} presents the distribution of similarity scores for each predicted class. The x-axis corresponds to the four categories (identical, useful, bad, incomprehensible), while the y-axis represents similarity scores. This visualization allows observation of how well classes produced by the model align with expected similarity levels, for example high scores for the identical class and lower scores for bad or incomprehensible classes. These results suggest that classification of speech recognition errors by LLMs constitutes a relevant indicator of system performance, while providing an interpretability gain compared to metrics based solely on cosine similarities.
 
\section{Conclusion}
 
This paper systematically evaluated the capabilities of large language models for automatic speech recognition evaluation through three complementary approaches. 
First, generative LLMs outperform classic and existing semantic metrics in the task of selecting the best hypothesis, with performance reaching 92-94\% agreement with human annotators, demonstrating effective capabilities for automatic annotation and for finding the best ASR system. Second, embeddings extracted from decoder-based LLMs provide high-quality representations, even with simple aggregation strategies. For this task, model size does not appear to be a determining factor, and models explicitly fine-tuned achieve the best performance across all pooling techniques. Third, LLMs appear capable of qualitatively classifying transcription errors in a manner consistent with human data, offering superior interpretability to opaque numerical metrics.
These results suggest that LLMs offer a promising and interpretable avenue for perceptual evaluation of ASR systems, paving the way for next-generation evaluation metrics that better align measured performance with actual human perception.



\section*{Acknowledgments}

This work was supported by the Idiap Research Institute and the ELOQUENCE project under the European Union’s Horizon 2020 programme (grant number 101070558)


\bibliography{custom}




\end{document}